\pdfoutput=1

\documentclass[11pt]{article}

\usepackage[]{emnlp2021}

\usepackage{times}
\usepackage{latexsym}
\usepackage{graphicx}
\usepackage{multirow}
\usepackage{enumitem}
\setlist{nosep}
\usepackage{amsfonts}
\usepackage[T1]{fontenc}

\usepackage[utf8]{inputenc}

\usepackage{microtype}

\title{Towards a Multi-modal, Multi-task Learning based Pre-training Framework for Document Representation Learning}

\author{Subhojeet Pramanik\thanks{$^{\star}$Authors contributed equally to the work} \\
  University of Alberta, Canada \\
  \texttt{spramanik@ualberta.ca} \\\And
  Shashank Mujumdar$^{\star}$ \\
  IBM Research, India \\
  \texttt{shamujum@in.ibm.com} \\ \And
  Hima Patel \\
  IBM Research, India \\
  \texttt{himapatel@in.ibm.com} \\
  }

\begin{document}
\maketitle
\begin{abstract}

Recent approaches in literature have exploited the multi-modal information in documents (text, layout, image) to serve specific downstream document tasks. However, they are limited by their - (i) inability to learn cross-modal representations across text, layout and image dimensions for documents and (ii) inability to process multi-page documents. Pre-training techniques have been shown in Natural Language Processing (NLP) domain to learn generic textual representations from large unlabelled datasets, applicable to various downstream NLP tasks. In this paper, we propose a multi-task learning-based framework that utilizes a combination of self-supervised and supervised pre-training tasks to learn a generic document representation applicable to various downstream document tasks. Specifically, we introduce \textit{Document Topic Modelling} and \textit{Document Shuffle Prediction} as novel pre-training tasks to learn rich image representations along with the text and layout representations for documents. We utilize the \textit{Longformer} network architecture as the backbone to encode the multi-modal information from multi-page documents in an end-to-end fashion. We showcase the applicability of our pre-training framework on a variety of different real-world document tasks such as document classification, document information extraction, and document retrieval. We evaluate our framework on different standard document datasets and conduct exhaustive experiments to compare performance against various ablations of our framework and state-of-the-art baselines.

\end{abstract}



\section{Introduction}

\subsection{Problem Motivation}

In the era of digitization, most businesses are turning towards leveraging artificial intelligence (AI) techniques to exploit the information contained in business documents. Traditional information extraction (IE) approaches utilize Natural Language Processing (NLP) methods to process the information from documents expressed in the form of natural language text \cite{manevitz2001one}. However, documents contain rich multi-modal information that includes both text and the document layout. The document layout organises the textual information into different formats such as sections, paragraphs, tables, multi-column etc. utilising different font-types/colors/positions/sizes/styles. Further, important visual cues are also indicated through figures/charts/logos etc. and the overall document page appearance. In general, information in a document spans over multiple pages which gives rise to a variety of complex document layouts that can be observed in scientific articles, invoices, receipts, emails, contracts, blogs, etc. Analyzing and understanding these documents is a challenging endeavor and requires a multi-disciplinary perspective combining NLP, computer vision (CV), and knowledge-representation to learn a generic document representation suitable for various downstream applications \cite{di}.

\begin{figure*}[t]
\centering
\includegraphics[width=0.7\textwidth]{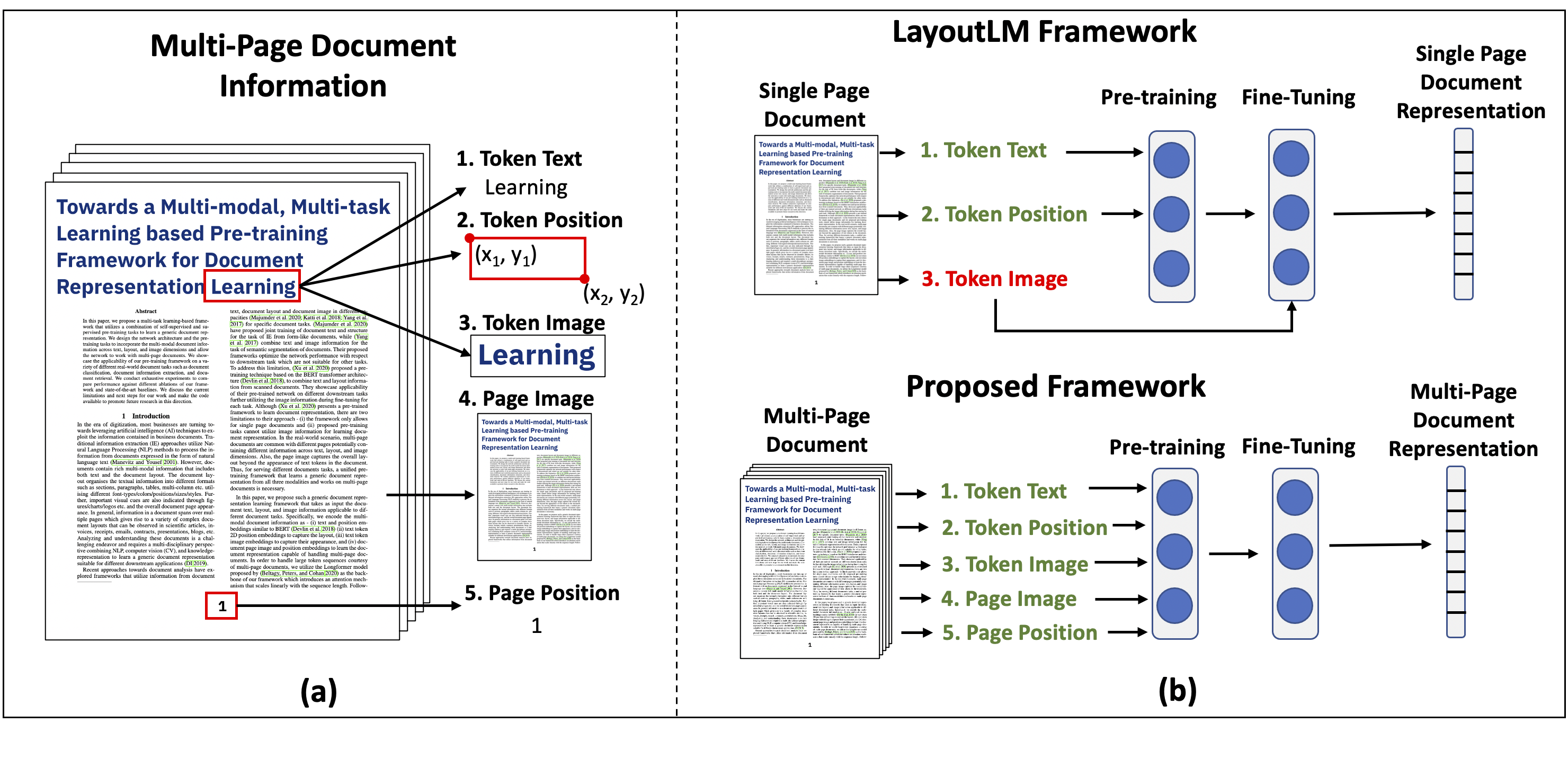}
\caption{(a) Multi-modal information used from input documents. (b) Distinction of proposed framework with LayoutLM \cite{xu2020layoutlm}}.
\label{fig:intro}
\end{figure*}

\begin{figure*}[t]
\centering
\includegraphics[width=0.98\textwidth]{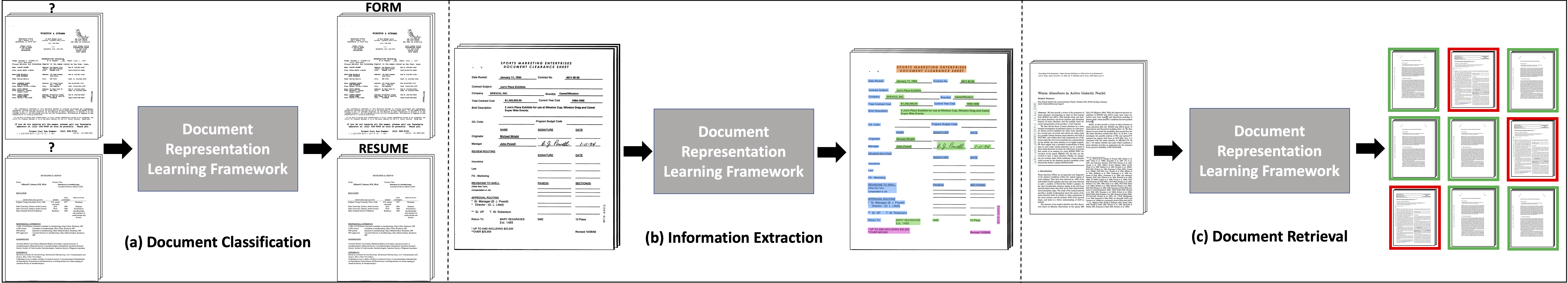}
\caption{Applicability of our framework on multi-page documents for different downstream tasks - (a) Document Classification (b) Information Extraction (c) Document Retrieval}.
\label{fig:applications}
\end{figure*}

\subsection{Limitations of Prior-art Approaches}

Recent approaches towards document analysis have explored frameworks that utilize information from document text, document layout and document image in different capacities \cite{majumder2020representation,katti2018chargrid,yang2017learning} for specific document tasks. In \cite{majumder2020representation}, the authors have proposed joint training of document text and structure for the task of IE from form-like documents, while in \cite{yang2017learning}, the authors combine text and image information for the task of semantic segmentation of documents. Their proposed frameworks optimize the network performance with respect to downstream task which are not suitable for other tasks. To address this limitation, in \cite{xu2020layoutlm} a pre-training technique is proposed based on the BERT transformer architecture \cite{devlin2018bert}, to combine text and layout information from a large set of unlabelled documents. They showcase applicability of their pre-trained network on different downstream tasks further utilizing the image information during fine-tuning for each downstream task. As shown in Figure \ref{fig:intro}(b), pre-training the network in an end-to-end fashion allows for cross-modality interaction which facilitates learning shared representations across document text tokens and their 2D positions/layout. However, there are two major limitations to their approach - (i) proposed pre-training tasks cannot utilize image information for learning the document representation and (ii) the framework only allows for single page documents. As shown in Figure \ref{fig:intro}(a), apart from the document text tokens and their positions, the visual appearance of the individual tokens as well as the overall page serve as an important indicator for learning the document representation. In the real-world scenario, multi-page documents are common with different pages potentially containing different information across text, layout, and image dimensions. Thus, for serving different documents tasks, a unified pre-training framework that learns a generic document representation from all three modalities and works on multi-page documents is necessary.

\subsection{Our Proposition}
In this paper, we propose a generic document representation learning framework that takes as input the document text, layout, and image information applicable to various document tasks as shown in Figure \ref{fig:intro}(a). Specifically, we encode the multi-modal document information as - (i) text and position embeddings similar to BERT \cite{devlin2018bert} (ii) text token 2D position embeddings to capture the layout, (iii) text token image embeddings to capture their appearance, and (iv) document page image and position embeddings to learn the document representation capable of handling multi-page documents. In order to handle large token sequences courtesy of multi-page documents, we utilize the Longformer model \cite{beltagy2020longformer} as the backbone of our framework which introduces an attention mechanism that scales linearly with the sequence length. We utilize the Masked Visual Language Modelling (MVLM) task and a document classification (CLF) task that enforces the joint pre-training of all the input embeddings \cite{xu2020layoutlm}. To facilitate the cross-modality interaction and ensure that the network learns from the visual information, we introduce two novel self-supervised pre-training tasks in our framework - (i) document topic modeling (DTM) and (ii) document shuffle prediction (DSP). We mine the latent topics from the document text and train our framework to predict the topic distribution using only the document page image embeddings for the task of DTM \cite{gomez2017self}. On the other hand, DSP involves shuffling the page image order while keeping the other embeddings intact for randomly sampled documents during training to identify if the document is tampered with. We employ a multi-task learning framework to simultaneously train multiple objectives of the different pre-training tasks to learn shared representations across the text, layout, and image modalities of the documents. Figure \ref{fig:applications} signifies the applicability of our pre-trained embeddings for various downstream document tasks. We evaluate the performance of our proposed framework on various document tasks - (i) Information Extraction (ii) Document Classification (iii) Table Token Classification and (iv) Document Retrieval. In summary, the main contributions of this work are:
\begin{itemize}[leftmargin=*]
    \item We introduce a multi-modal, multi-task learning based pre-training framework to learn a generic document representation.
    \item We introduce document topic-modeling and document shuffle prediction as self-supervised pre-training tasks
    \item Our proposed framework is able to process multi-page documents which is a limitation of prior art approaches
    \item We conduct exhaustive experiments to compare performance against various ablations of our framework and SOTA baselines
\end{itemize}

\section{Approach}
In this section, we describe the details of our proposed architecture. 

\begin{figure*}[t]
\centering
\includegraphics[width=0.9\linewidth]{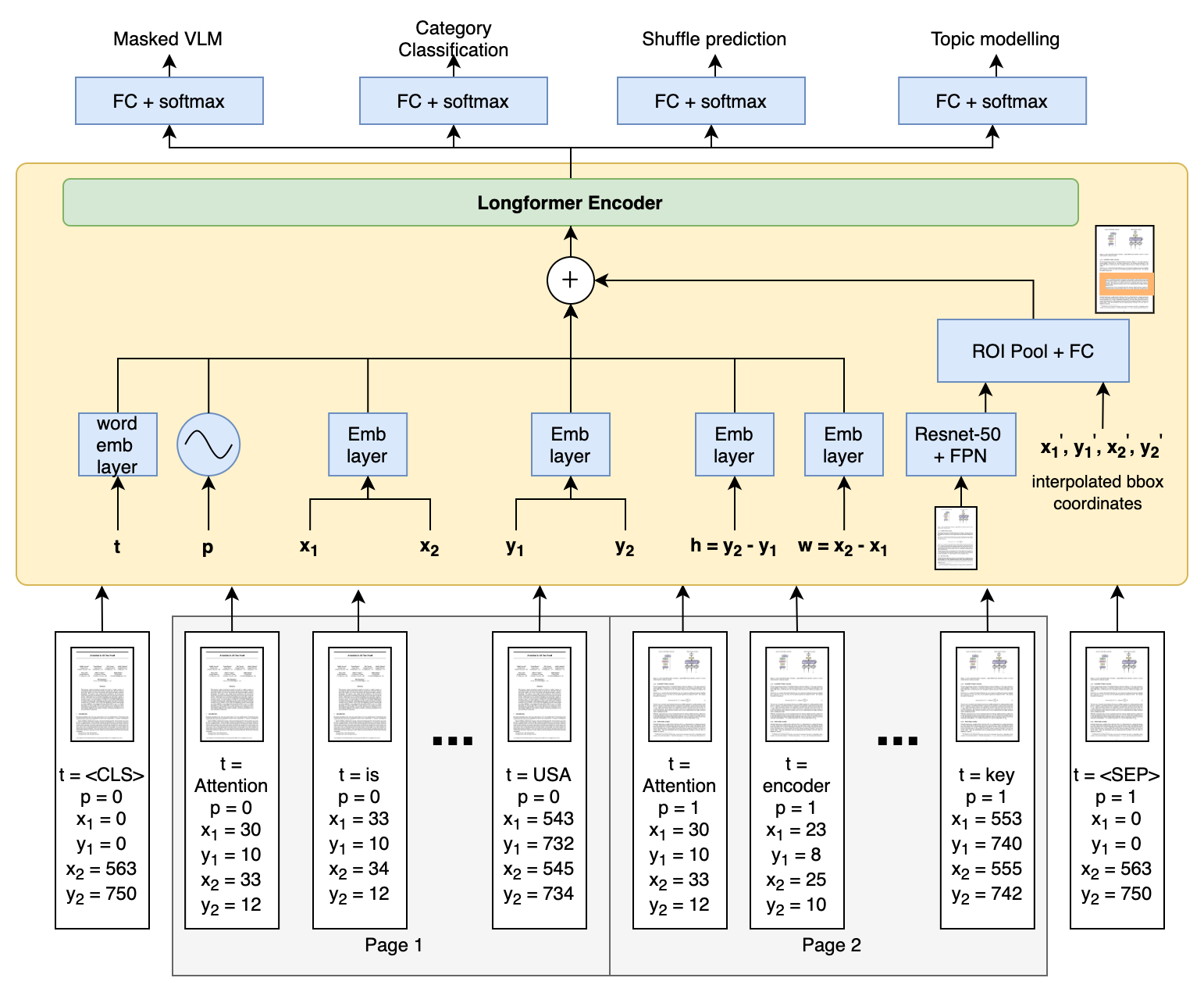} 
\caption{Demonstration of the proposed architecture encoding two sample pages from a PDF document. }

\label{fig:framework}
\end{figure*}

\subsection{The Proposed architecture}

Common Transformer variants such as BERT \cite{devlin2018bert} \& RoBERTa \cite{Roberta} are adept at processing text tokens and learning semantic representations from text sequences. However, they do not leverage the visual and layout information present in documents. LayoutLM \cite{xu2020layoutlm}, incorporates the layout information provided by the token level 2D bounding boxes and adds these layout embeddings to the existing token level embeddings in BERT architecture during pre-training. We introduce two new embeddings in addition to layout \& token embeddings: (1) the visual information present in the corresponding bounding box for each token, and (2) the page-level information present in multi-page PDF documents (as shown in Figure \ref{fig:intro}. Additionally, we use the Longformer \cite{beltagy2020longformer} network architecture for encoding the multi-modal inputs.
BERT variants \cite{devlin2018bert} scale quadratically in terms of memory and CPU requirements, \cite{beltagy2020longformer} introduces local windowed attention and task-motivated global attention mechanisms that scales linearly with sequence length.  

Figure \ref{fig:framework} showcases our proposed network architecture. For a multi-page document we parse its constituent tokens using standard document parser \cite{pytesseract} and store the token text and the corresponding bounding boxes along with the page numbers and page images. Every document is encoded as a sequence of tokens $t \in \mathbb{Z}, 0\leq t < n_v$, page numbers $p \in \mathbb{Z}, 0 \leq p < n_p$, bounding box coordinates $(x_1,y_1,x_2,y_2)$, and the image of the entire page corresponding to the given token; where $n_v$ is the vocab size, $n_p$ is the maximum number of page embeddings; $(x_1,y_1)$ are the coordinates for the upper left and $(x_2,y_2)$ are the coordinates for the lower right corner of the bounding box, and $h=y_2-y_1$, $w=x_2-x_1$ capture the height and width of the bounding box respectively. We use four embedding layers to encode the layout information: X dimension ($x_1, x_2$), Y dimension ($y_1, y_2$), $h$ and $w$. Embeddings from the same dimension share the embedding layers. 

Novel to our approach, we use a ResNet-50 \cite{resnet} architecture combined with Feature Pyramid Network (FPN) \cite{fpn} to generate multi-level image embeddings for the given page corresponding to each token. For an image of size $(u,v)$, the Resnet+FPN layer produces feature maps of size $(u^{'},v^{'})$. The bounding boxes $(x_1,y_1,x_2,y_2)$ which are originally in range $x \in \mathbb{Z}, 0\leq x\leq u$ \& $y \in \mathbb{Z}, 0\leq y \leq v$ are linearly scaled to match the feature map dimension $(u^{'},v^{'})$ respectively. We select the final layer of the FPN network which has the highest semantic representation. To generate the final image embedding corresponding to the region indicated by the bounding box coordinates, a Region of Interest (RoI) pooling operation \cite{roipool} is performed on the page image feature map with an output size of $1\times 1$ using the interpolated bounding box coordinates. 
Using RoI pooling allows us to efficiently select the embeddings for the regions indicated by all the bounding boxes for each token from the page feature map. For each token, we also pass the corresponding page number through an embedding layer initialized using sinusoidal embeddings, as described in \cite{transformer}. The embeddings for the images, layout, and pages are added to the existing text embeddings and passed to the Longformer encoder to generate sequence representations for the document.


For the special tokens $<$CLS$>$ and $<$SEP$>$ which are predominantly used in BERT variants for sequence inputs, we use $(0,0,u,v)$ as the bounding box as it captures the image embedding for the entire page, thereby benefitting the downstream tasks that require the representation of the $<$CLS$>$ token for prediction.  For, all our experiments, we freeze all except the last layer of Resnet-50.

\subsection{Multi-task learning framework}
We use a multi-task learning framework to pre-train our network on a combination of three self-supervised tasks that are posed as classification tasks along with a supervised category classification task. At each training step, we optimize all the pre-training tasks in a joint fashion. For each pre-training task, the task-specific inputs are encoded according to their respective input strategies, and the task-specific loss is calculated. The gradients are computed with respect to each task-specific loss and accumulated across all tasks to be optimized using the AdamW optimizer \cite{adamw}. All tasks use cross-entropy loss for classification except DTM which uses soft cross-entropy loss.



\subsection{Pre-training dataset:} 
We use Arxiv PDFs \cite{arxiv} for pre-training our architecture comprising of scientific articles belonging to 16 different categories such as mathematics, physics, computer science, etc. 
We extract the first 130k PDF documents from the arxiv bulk server and use a (train, val, test) split of (110k, 10k, 10k) respectively. 
We process the documents \cite{pytesseract} to extract and store the text tokens, corresponding bounding boxes, page numbers, and the page images along with the document category to feed as input to our network.



\subsection{Pre-training tasks:} 


\textbf{1. Masked Visual Language Modelling (MVLM):}
BERT model utilizes Masked Language Modelling (MLM) where input tokens are masked during pre-training and predicted as output using the context from non-masked tokens. Compared to MLM, in \cite{xu2020layoutlm} MVLM is introduced which masks the input tokens by replacing it with a designated  $<$MASK$>$ token, but keeps only the layout information provided by the bounding boxes. It however does not utilize the visual information of the tokens during pre-training. On the other hand, we also utilize the image embedding generated by the Resnet+FPN layers along with the layout embeddings at the masked locations. 


\textbf{2. Document Category Classification (CLF):} 
Each document in Arxiv dataset belongs to one of 16 categories denoting the relevant subject area of the document. The category prediction is performed at the $<$CLS$>$ token, by passing the output of token into a fully-connected (FC) layer appended with a softmax classification layer.

\textbf{3. Document Shuffle Prediction (DSP):} 
For DSP, given a document, we randomly shuffle the order of the page images while preserving the order of other embeddings before passing to the network. Thus, although the token text and bounding box embeddings are in order, the corresponding token image embeddings are uncorrelated since the page images are shuffled. For a given document, the page images are shuffled with a probability of 0.5, and the model is trained to predict whether the input document pages are shuffled or intact using all the embeddings. We argue that, in order to successfully train on the DSP task, the network is forced to correlate the token image embeddings with the corresponding token text and bounding box embeddings. 


\textbf{4. Document Topic Modelling (DTM):}
Although training on the DSP task enforces the network to correlate image and text modalities at the token level, we introduce the DTM task to learn improved page image representations.
The objective is to learn discriminative visual features employing the semantic context as soft-labels during training \cite{gomez2017self}. We encode the semantic context for each document as a probability distribution over a set of latent topics. We utilize the Latent Dirichlet Allocation (LDA) algorithm \cite{blei2003latent} to mine the latent topics over the set of text tokens parsed from the Arxiv training set. During training, the vector of topic probabilities is computed using the learned LDA model for each document. For the DTM task, we pass the page images of the document to our network, while a single $<$MASK$>$ token is passed for text embedding. Further, the bounding box coordinates of the complete page are passed as part of the layout embedding. A soft cross-entropy loss is applied to the predicted output of the network against the vector of topic probabilities for learning. Since the Arxiv dataset has 16 subject areas as categories within the documents, we chose to mine 30 latent topics to further identify granular categorization among the documents.

\begin{table*}[t]
    \caption{Model performance numbers for the Arxiv Classification task. (Prec: Precision, Rec: Recall)}
    \label{table:arxiv}
    \centering
    \begin{tabular}{lllccccc}
    \hline
    \hline
    Model & Input & Pre-training Tasks & Pre-training Size & Prec & Rec & F1 \\
    \hline
    Our & Text & MLM+CLF & 110K (5 epochs) & 90.72\% & 90.40\% & 90.46\%\\
    Our & Text+Layout & MVLM+CLF & 110K (5 epochs)  & 90.79\% & 90.72\% & 90.71\% \\
    Our & All  & MVLM+CLF & 110K (5 epochs)  & 98.92\% & 98.90\% & 98.90\%\\
    Our & All  & MVLM+CLF+DSP & 110K (5 epochs) & 98.91\% & 98.90\% & 98.90\%\\
    Our & All  & MVLM+CLF+DTM & 110K (5 epochs) & \underline{98.92\%} & \underline{98.91\%} & \underline{98.92\%}\\
    Our & All & All & 110K (5 epochs) & \textbf{98.93}\%  & \textbf{98.92\%} & \textbf{98.93\%}\\
    \hline
    \hline
    \end{tabular}
\end{table*}

\begin{table*}[t]
    \caption{Model performance numbers for the RVL-CDIP classification task.  LayoutLM\textsuperscript{*}\textsubscript{BASE} uses Resnet-101 image embeddings during fine-tuning. (Acc: Accuracy)}
    \label{table:rvlcdip}
    \centering
    \begin{tabular}{p{5.9cm}llcp{1cm}}
    \hline
    \hline
    Model & Input & Pre-train Tasks & Pre-train Size & Acc  \\
    \hline
    
    Our & Text & MLM+CLF & 110K (5 epochs) & 84.48\% \\
    Our & Text+Layout & MVLM+CLF & 110K (5 epochs) & 86.55\%  \\
    Our & All & MVLM+CLF & 110K (5 epochs) & 91.22\% \\
    Our & All & All  & 110K (5 epochs) & 91.72\% \\
    Our (VGG-16) & All & All  & 110K (5 epochs) & \underline{93.36\%} \\
    \hline
    LayoutLM\textsubscript{BASE} & Text+Layout & MVLM & 500K (6 epochs) & 91.25\%  \\
    LayoutLM\textsuperscript{*}\textsubscript{BASE} & Text+Layout & MVLM+MDC &1M (6 epochs)  & \textbf{94.31\%} \\
    VGG-16 \citeauthor{afzal2017cutting} & Image & - & -&  90.97\% \\
    Stacked CNN Ensemble \citeauthor{das2018document} & Image & - & - & 92.21\% \\
    LadderNet \citeauthor{sarkhel2019deterministic} & Image & - & -& 92.77\% \\
    Multimodal Ensemble \citeauthor{dauphinee2019modular} & Text+Image & - & - & 93.07\% \\
    \hline
    \hline
    \end{tabular}
\end{table*}

\section{Datasets and Experiments}
\subsection{Datasets}

\textbf{FUNSD:} The FUNSD dataset \cite{jaume2019} consists of 199 fully annotated, scanned single-page forms with overall 31,485 words. Semantic entities comprising of multiple tokens are annotated with labels among `question', `answer', `header', or `other'. Additionally, the text, bounding boxes for each token, and links to other entities are present. The dataset has 149 train and 50 test images. We evaluate our network on the semantic labeling task and measure the word-level F1 scores \cite{xu2020layoutlm}. 

\textbf{RVL-CDIP:} The RVL-CDIP dataset \cite{harley2015icdar} consists of 400,000 grayscale images organized into 16 classes, with around 25,000 images per class. The images are characterized by low quality, presence of noise, and low resolution, typically 100 dpi. The dataset consists of 320,000 training, 40,000 validation and 40,000 test images.  The 16 classes include {letter, form, email, handwritten, advertisement, scientific report, scientific publication, specification, file folder, news article, budget, invoice, presentation, questionnaire, resume, memo}. We evaluate our architecture on the document classification task using the 16 labels.  

\textbf{ICDAR19:} The Track A Modern data released as part of the ICDAR19 dataset \cite{gao2019icdar} contains 600 train \& 240 test images from various PDF documents such as scientific journals, forms, financial statements, etc. annotated with table bounding box coordinates. We perform word-level binary classification on this dataset.

\subsection{Model Pre-training}

We initialize the Longformer Encoder and Word embedding layer with the pre-trained weights from the Longformer\textsubscript{BASE} (12 layers, 512 hidden size) \cite{beltagy2020longformer}, as shown in Figure \ref{fig:framework}. We utilize a global+sliding window attention of length 512. The weights of Resnet-50 are initialized using the Resnet-50 model pre-trained on the ImageNet dataset \cite{resnet}. Across all pre-training and downstream tasks, we resize all page images to $563\times 750$ and correspondingly scale the bounding box coordinates. We limit the maximum number of pages to 5 per document and limit the number of tokens to 500 per page during pre-training for sequence classification tasks. For the different pre-training tasks, we use a batch size (BS) and gradient accumulation (GA) of - (i) MVLM \& CLF (BS=32 \& GA=2); (ii) DSP (BS=16 \& GA=1); (iii) DTM (BS=16 \& GA=1). We pre-train our architecture for 15K steps ($\sim$5 epochs) with a learning rate of 3e-5 on a single NVIDIA Tesla V100 32GB GPU.

\subsection{Experiment Setup}
We evaluate our model on the following different downstream tasks to demonstrate its efficacy. 

\textbf{Document Classification:} 
We finetune our model to perform multi-page document classification on the Arxiv dataset and single-page document classification on the RVL-CDIP dataset. For both tasks, each document is encoded as a sequence of tokens, bounding boxes, page images, page numbers, as shown in Figure \ref{fig:framework}. We use \cite{pytesseract} to extract the word-level tokens and bounding boxes. The category prediction is performed at the $<$CLS$>$ token by passing its output through an FC+Softmax layer. We use a learning rate of 3e-5, (BS=12 \& GA=4) for Arxiv and (BS=64 \& GA=1) for RVL-CDIP, and we fine-tune our model and different ablations for 5 epochs for both datasets independently. We use weighted precision, recall as F1 as our evaluation metric. 

\textbf{Form Understanding:} 
We perform the semantic labeling task on the FUNSD dataset as a sequence labeling problem. Each form is treated as a single-page document and sequenced as a list of tokens, bounding boxes, and the page image. For each token, we pass its learned representation through an FC+Softmax layer to predict its category. We use word-level weighted precision, recall, and F1 score as the evaluation metrics \cite{xu2020layoutlm}. For fine-tuning, we use BS=12, GA=1, a learning rate of 3e-5, and train for 20 epochs. 

\textbf{Table Token Classification:} 
For this task, the model is fine-tuned as a sequence labeling problem to classify the word-tokens in a document as `table' or `other'. Table bounding boxes are used to generate ground truth labels for each token in the document as detected using \cite{pytesseract}. Processing the document, generating the input embeddings, and the token level prediction is performed similarly to the Form Understanding task. For fine-tuning, we use BS=4, GA=2, a learning rate of 3e-5, and train for 14 epochs.

\textbf{Document Retrieval:} 
Similar to the classification task, we process the multi-page documents in the Arxiv dataset and fine-tune our pre-trained model with all inputs on all pre-training tasks and the BERT\textsubscript{BASE} model with only the text input from the Arxiv training set. We utilize 10k documents from the Arxiv test set split into 2k query and 8k retrieval set. For a given query document, we use the fine-tuned embeddings from each model and compute its cosine distance with the query set for retrieval. We compare the mean average precision (MAP) and the normalized discounted cumulative gain (NDCG) for evaluation.

\begin{table*}[t]
    \caption{Model performance numbers for the semantic labelling task on FUNSD dataset. LayoutLM\textsuperscript{*}\textsubscript{BASE} uses Resnet-101 image embeddings during fine-tuning. (Prec: Precision, Rec: Recall)}
    \label{table:funsd}
    \centering
    \begin{tabular}{lllcp{1cm}p{1cm}p{1cm}}
    \hline
    \hline
    Model & Input & Pre-train Tasks& Pre-train Size & Prec & Rec & F1 \\
    \hline
    Our & Text & MLM+CLF & 110K (5 epochs) & 77.25\% & 68.40\% & 69.66\% \\
    Our & Text+Layout & MVLM+CLF & 110K (5 epochs)  & 75.45\% & 74.93\% & 75.15\% \\
    Our & All & MVLM+CLF & 110K (5 epochs)  & 77.31\% & 76.50\% & 76.79\%\\
    Our & All & MVLM+CLF+DSP& 110K (5 epochs) & 77.55\% &76.80\% & 77.30\%\\
    Our & All & MVLM+CLF+DTM& 110K (5 epochs) & \underline{77.84\%} & 77.10\% & \underline{77.42\%}\\
    Our & All & All & 110K (5 epochs) & \textbf{78.41\%} & \underline{77.35\%} & \textbf{77.44\%} \\
    \hline
    LayoutLM\textsubscript{BASE} & Text+Layout & MVLM & 500K (6 epochs)  & 66.50\% & 73.55\% & 69.85\% \\
    LayoutLM\textsuperscript{*}\textsubscript{BASE} & Text+Layout & MVLM+CLF & 1M (6 epochs) & 71.01\% & \textbf{78.15\%} & 74.41\% \\
    \hline
    \hline
    \end{tabular}
\end{table*}

\begin{table}[t]
    \caption{Results of Multi-page Document Retrieval Task}
    \label{table:retrieval}
    \centering
    \begin{tabular}{l|ccc}
    \hline
    \hline
    Model & MAP & NDCG-1 & NDCG-10 \\ \hline
    BERT & 91.01\% & 90.08\% & 93.00\% \\
    Our\textsubscript{All} & \textbf{98.99\%} & \textbf{98.94\%} & \textbf{99.21\%} \\
    \hline
    \hline
    \end{tabular}
\end{table}


\section{Results and Discussion}
\begin{table}[t]
    \caption{Inference Ablation Results on the FUNSD dataset (F1 score)}
    \label{table:inference_ablation}
    \centering
    \begin{tabular}{p{3cm}|p{1cm}p{1cm}p{1cm}}
    \hline
    \hline
    \multirow{2}{2.6cm}{Pre-training Tasks} & \multicolumn{3}{c}{Inference Ablations} \\ \cline{2-4}
    & All & Text\textsubscript{Only} & Image\textsubscript{Only} \\ \hline 
    MVLM+CLF & 76.79\% & 73.64\% & 33.24\% \\
    MVLM+CLF+DSP& 77.30\% & 74.10\% & 35.42\% \\
    MVLM+CLF+DTM& 77.42\% & 74.68\% & 38.20\% \\
    All & \textbf{77.44\%} & \textbf{75.10\%} & \textbf{40.12\%} \\
    \hline
    \hline
    \end{tabular}
\end{table}

\textbf{Multi-page Document Classification: }
Prior-art approaches do not support multiple-page documents as their network architecture does not enable to encode multi-page information. Thus, we compare the results of various ablations of our architecture for this task on the Arxiv dataset. As seen in Table \ref{table:arxiv}, a significant boost in performance can be observed with the introduction of the image embeddings to the pre-training against the Text and Text+Layout ablations. The DSP and DTM tasks further improve the performance marginally. The ablation involving the DTM pre-training task showcases higher improvement which suggests that utilizing the image information to predict the topic distribution during pre-training helps in learning improved image embeddings. The performance gain is also attributed to the underlying sequence encoder Longformer\textsubscript{BASE} with an attention mechanism up to 4096 tokens which enable multi-page processing to learn multi-modal contextual representations.

\textbf{Single page Document Classification:}
Table \ref{table:rvlcdip} shows the results of the single-page document classification on the RVL-CDIP dataset. All our models are pre-trained on the Arxiv dataset, LayoutLM is pre-trained on the IIT-CDIP dataset, whereas all other baselines perform no pre-training. Although our standard model with Resnet image layers beats the comparable LayoutLM\textsubscript{BASE} model pre-trained on 500K documents and image-based VGG-16 model, the task-specific approaches such as Stacked CNN Ensemble, LadderNet, and Multimodal Ensemble outperform our model. Since RVL-CDIP inherently contains low-quality images, task-specific approaches propose clever network architectures to utilize discriminative multi-scale features \cite{sarkhel2019deterministic}, multiple VGG-16 models to process different parts of document image \cite{das2018document} and augmenting image features with raw text features \cite{dauphinee2019modular} to achieve high classification performance. Further, it is known that Resnet-50 performs poorly on the RVL-CDIP dataset and even a smaller network such as VGG-16 performances better \cite{afzal2017cutting}. Thus, we also consider a variation of our architecture with the Resnet-50 image layers replaced with the VGG-16 image layers. With the VGG-16 layers, we see a significant improvement in the performance from 91.72\% to 93.36\% and our VGG-16 based model beats the existing image-based and image+text-based baselines trained on comparable dataset sizes. We achieve comparable performance with fine-tuning on 5 epochs to LayoutLM\textsuperscript{*}\textsubscript{BASE} which is pre-trained on 1M documents and fine-tuned using the Faster R-CNN embeddings for 30 epochs.


\textbf{Semantic Labelling Task: }
We present results of our model fine-tuned on FUNSD semantic labeling task in Table \ref{table:funsd}. 
Our best model pre-trained on all four tasks and all inputs achieves an F1 score of 77.44\% outperforming the comparable LayoutLM\textsuperscript{*}\textsubscript{BASE} model which achieves an F1 score of 74.41\%. We attribute the increase in performance to the inclusion of RoI pooled image embeddings indicated by the various bounding box regions for each text token during pre-training as well as fine-tuning the Resnet-50 layers and the LayoutLM architecture is agnostic to both these properties.
Further, our architecture pre-trains using 110K documents compared to LayoutLM which uses 500K \& 1M documents. Thus, we argue that even for a significantly smaller dataset size, our model generalizes better by incorporating image embeddings during pre-training on the FUNSD task. 

\begin{figure}[t]
\centering
\includegraphics[width=0.99\linewidth]{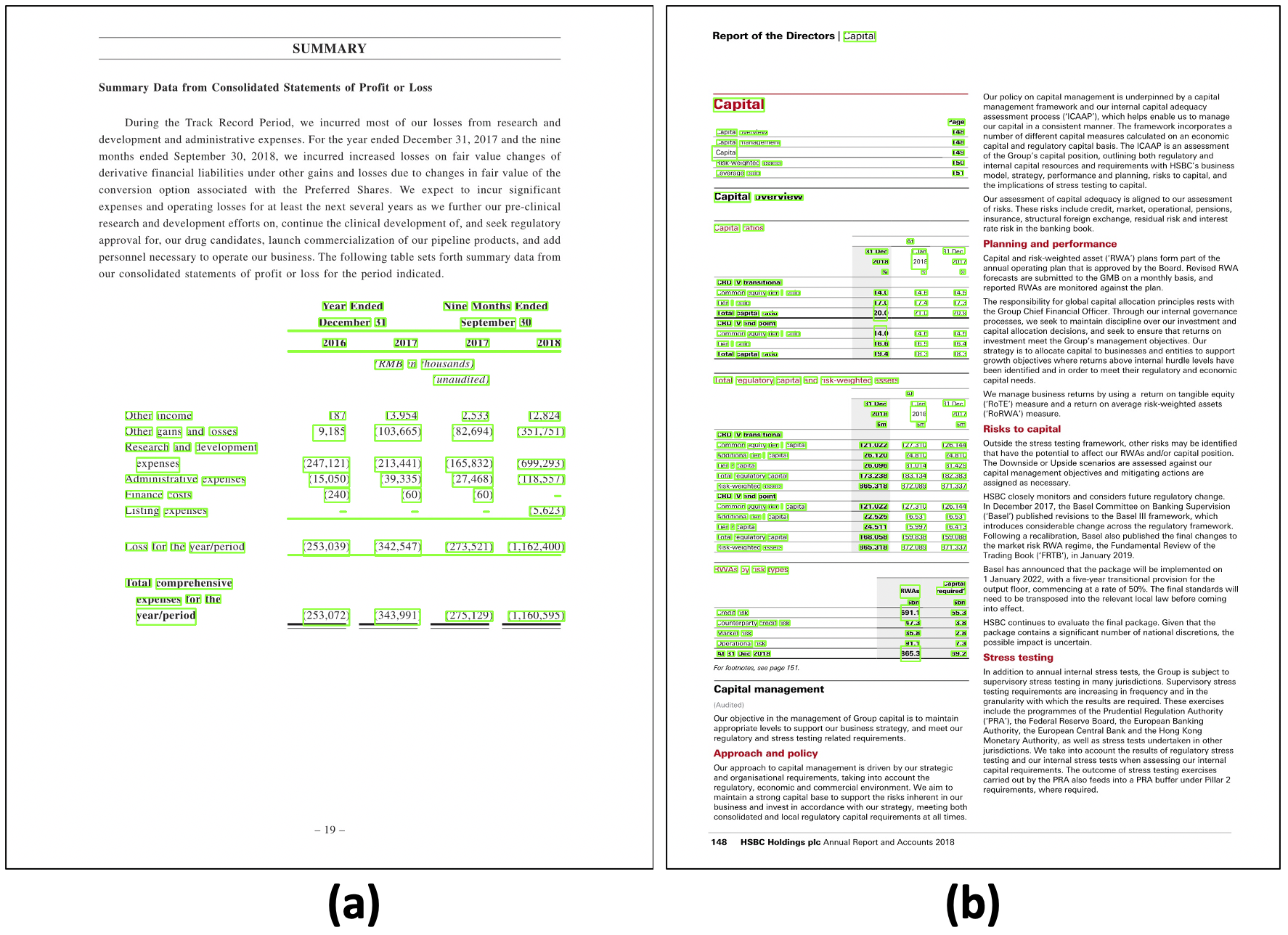} 
\caption{Results of Table Token Classification. Tokens predicted as ``table" by our model are marked in green.}.
\label{fig:table_token}
\end{figure}

Similar to the FUNSD task, the table token classification task performs semantic labeling. However, the impact of jointly learning the text, layout, and image embeddings is much more evident from the results shown in Figure \ref{fig:table_token}. Our model can correctly classify all the tokens belonging to tables with a negligible amount of false positives. We get the precision, recall, and f1-score of 94.99\%, 94.98\%, and 94.97\% respectively on the ICDAR2019 test set. It is noteworthy that only fine-tuning our model on the train set can achieve promising results, which the prior art approaches employ careful heuristics to achieve \cite{gao2019icdar}.

\textbf{Multi-page Document Retrieval Task: }
To ascertain the utility of our proposed framework to process multi-page documents, we compare the results of our framework against the standard BERT model for the task of multi-page document retrieval in Table \ref{table:retrieval}. Fine-tuned embeddings from our model significantly outperform those from the BERT model. The high value of MAP and NDCG-10 indicates that the retrieved samples are not only correct but also ranked higher than the incorrect ones for most of the queries. Although our model captures richer embeddings, the significant boost in performance is also attributed to the Longformer architecture that can encode much more information across document pages compared to vanilla BERT architecture.

\subsection{Inference Ablation Study}
In order to further investigate the utility of DSP and DTM tasks, we compare the performance of four models during inference on the FUNSD task 
For each model, we conduct two ablations during inference, where only the text or image embedding is used to make the prediction while excluding the layout embeddings. As shown in Table \ref{table:inference_ablation}, for both the ablations, Our\textsubscript{DSP}, Our\textsubscript{DTM} \& Our\textsubscript{All} retains higher performance than Our\textsubscript{MC}. 
In particular, for the image-only ablation, the difference in the performance drop (All\% - Image Only\%) for Our\textsubscript{All} ($\sim$37\%) is lower than that for Our\textsubscript{MC} ($\sim$43\%). Similarly, for Our\textsubscript{DSP} ($\sim$42\%) and Our\textsubscript{DTM} ($\sim$39\%), the performance drop is lower than that for Our\textsubscript{MC}, however, higher than Our\textsubscript{All}. This suggests that with the introduction of DSP and DTM tasks, the learnt image embeddings for Our\textsubscript{DSP}, Our\textsubscript{DTM} \& Our\textsubscript{All} capture richer image representations than Our\textsubscript{MC}. 
These pre-training tasks help learn better image representations that retain more performance even when the text information is missing during inference. The trend observed is similar when using text only embeddings. However, the difference is not that significant as both share MVLM and classification tasks which are more adept at learning textual representations.

\section{Related Work}

In recent years, different prior-art approaches have focussed on taking a multi-modal approach for document understanding. Exploring the document semantics as well as structure allow to learn a granular understanding of the document information necessary to solve problems such as information extraction, semantic segmentation, layout analysis, table structure detection etc. These approaches fundamentally involve analysing document structure in addition to the primary modality of document text or document image. \cite{katti2018chargrid} introduce a document representation that encodes the character level textual information while preserving the 2D document layout. They train a fully convolutional encoder-decoder network that learns from this input representation to extract semantic information from invoices. For similar task of information extraction from invoices, \cite{zhao2019cutie} propose a convolutional network that learns both semantic and structural information from scanned invoices by creating a gridded text representation that preserves the spatial relationship among the text tokens. Contrary to these approaches, \cite{majumder2020representation} utilize the knowledge of key fields to be extracted from a document to generate candidates and learn their dense representation that encodes information from its positional neighbors. For analysing the tables in scanned documents, \cite{schreiber2017deepdesrt,paliwal2019tablenet,prasad2020cascadetabnet} propose different modifications to standard CNN network architectures such as VGGNet \cite{simonyan2014very} used for classification and Faster R-CNN \cite{ren2015faster} for object detection in images to recognise tables and identify their structure. Similarly, \cite{soto2019visual} propose to augment the Faster R-CNN object detection network architecture \cite{ren2015faster} with contextual features about the document pages and region bounding boxes to segment key regions in scientific articles. On the contrary, \cite{yang2017learning} propose to solve this as a pixel-wise semantic segmentation task utilising a multi-modal encoder-decoder network architecture that takes as input both the text and image embeddings. To learn a generic representation for supporting different tasks such as document image classification and document information extraction, \cite{xu2020layoutlm} propose to utilise the BERT transformer architecture \cite{devlin2018bert} to encode text as well as layout information to learn pre-trained embeddings and further utilise image information to fine-tune for a specific task. 

Most of the approaches in prior art utilize the multi-modal document information from single-page documents and extending their applicability to multi-page documents needs further exploration. Further, these approaches rely on limited labeled data, thus, exploring self-supervised learning to leverage large unlabeled datasets also needs exploration. We attempt to address these limitations in this paper.


\section{Conclusion and Future work}
We present a multi-modal pre-training framework that utilizes multi-task learning to learn a generic document representation. Our framework encodes the visual, layout, and textual information and supports real-world multi-page documents. Our network is pre-trained on the publically available Arxiv dataset utilizing self-supervised tasks that promote learning multi-modal shared representations across the text, layout, and image dimensions. We fine-tune our pre-trained network to showcase state-of-the-art performance on different document tasks such as document classification, information extraction, and document retrieval. In future, we will investigate pre-training on large datasets such as PubLayNet \cite{zhong2019publaynet} 
and further explore new architecture designs that will enable document image tasks such as table detection/page segmentation using our framework.

\bibliographystyle{acl_natbib}
\bibliography{main}

\end{document}